\documentclass[acmtog,authorversion]{acmart}
\acmSubmissionID{1483}

\usepackage[ruled]{algorithm2e} % For algorithms

\SetAlFnt{\small}
\SetAlCapFnt{\small}
\SetAlCapNameFnt{\small}
\SetAlCapHSkip{0pt}
\usepackage{bm}

\newcommand{\method}{PanoDreamer}

\newcommand{\edit}[1]{{#1}}
\usepackage{amsmath}

\DeclareMathOperator*{\argmin}{arg\,min}

\DeclareRobustCommand
  \Compactdots{\mathinner{\cdotp\mkern-2mu\cdotp\mkern-2mu\cdotp}}

\newcommand{\PanoMethod}{MultiConDiffusion}
\newcommand{\DepthMethod}{PanoDepthFusion}

\copyrightyear{2025}
\acmYear{2025}
\setcopyright{cc}
\setcctype{by}
\acmConference[SA Conference Papers '25]{SIGGRAPH Asia 2025 Conference Papers}{December 15--18, 2025}{Hong Kong, Hong Kong}
\acmBooktitle{SIGGRAPH Asia 2025 Conference Papers (SA Conference Papers '25), December 15--18, 2025, Hong Kong, Hong Kong}
\acmDOI{10.1145/3757377.3763883}
\acmISBN{979-8-4007-2137-3/2025/12}

\begin{document}
% Title portion
\title{PanoDreamer: Optimization-Based Single Image to 360 3D Scene With Diffusion}

% DO NOT ENTER AUTHOR INFORMATION FOR ANONYMOUS TECHNICAL PAPER SUBMISSIONS TO SIGGRAPH 2019!
\author{Avinash Paliwal}
\orcid{1234-5678-9012-3456}
\affiliation{%
 \institution{Texas A\&M University}
 \city{College Station}
 \state{TX}
 \country{USA}}
\affiliation{%
 \institution{Morphic Inc}
 \city{San Jose}
 \state{CA}
 \country{USA}}
\email{avinashpaliwal@tamu.edu}
\author{Xilong Zhou}
\affiliation{%
 \institution{Max Planck Institute for Informatics}
 \city{Saarbrücken}
 % \state{TX}
 \country{Germany}}
\email{xzhou@mpi-inf.mpg.de}
\author{Andrii Tsarov}
\affiliation{%
\institution{Leia Inc.}
\city{Mountain View}
\state{CA}
\country{USA}}
\email{andrii.tsarov@leiainc.com}
\author{Nima Kalantari}
\affiliation{%
 \institution{Texas A\&M University}
 \city{College Station}
 \state{TX}
 \country{USA}}
\email{nimak@tamu.edu}

\begin{abstract}
In this paper, we present \method{}, a novel method for producing a coherent 360\textdegree{} 3D scene from a single input image. Unlike existing methods that generate the scene sequentially, we frame the problem as single-image panorama and depth estimation. Once the coherent panoramic image and its corresponding depth are obtained, the scene can be reconstructed by inpainting the small occluded regions and projecting them into 3D space. Our key contribution is formulating single-image panorama and depth estimation as two optimization tasks and introducing alternating minimization strategies to effectively solve their objectives. We demonstrate that our approach outperforms existing techniques in single-image 360\textdegree{} 3D scene reconstruction in terms of consistency and overall quality\footnote{{\url{people.engr.tamu.edu/nimak/Papers/PanoDreamer}}}.

\end{abstract}

\begin{CCSXML}
<ccs2012>
   <concept>
       <concept_id>10010147.10010371.10010382.10010385</concept_id>
       <concept_desc>Computing methodologies~Image-based rendering</concept_desc>
       <concept_significance>500</concept_significance>
       </concept>
 </ccs2012>
\end{CCSXML}
\ccsdesc[500]{Computing methodologies~Image-based rendering}


\keywords{Single Image to 3D, 3D Scene Generation,
Diffusion Models, Panorama Generation, Panorama Depth}

\begin{teaserfigure}
\centering
  \includegraphics[width=0.85\textwidth]{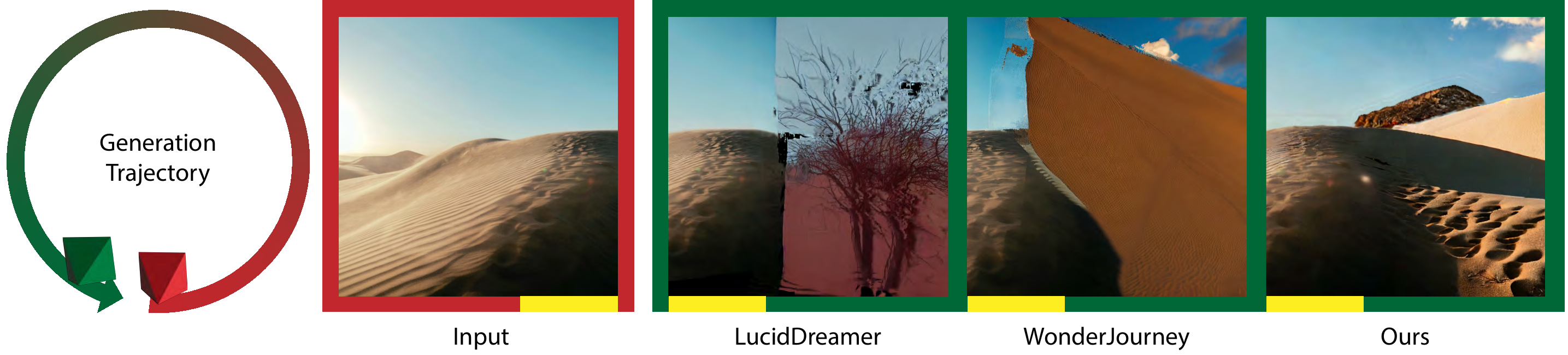}
  \vspace{-0.15in}
  \caption{We introduce a novel method for 360\textdegree{} 3D scene synthesis from a single image. Our approach generates a panorama and its corresponding depth in a coherent manner, addressing limitations in existing state-of-the-art methods such as LucidDreamer~\cite{chung2023luciddreamer} and WonderJourney~\cite{yu2024wonderjourney}. These methods sequentially add details by following a generation trajectory, often resulting in visible seams when looping back to the input image. In contrast, our approach ensures consistency throughout the entire 360\textdegree{} scene. The yellow bars show the regions corresponding to the input in each result.}
  \label{fig:teaser}
  \Description{}
\end{teaserfigure}

\maketitle

\section{Introduction}
\label{sec:intro}

Generating immersive and realistic 3D scenes from a single input image has emerged as one of the important topics in computer vision/graphics, driven by its broad applications including virtual/augmented reality (VR/AR) and gaming. While early algorithms~\cite{niklaus20193d, shih20203d, kopf2020one, jampani2021slide, zhou2016view, srinivasan2017learning, li2020synthesizing, tucker2020single} have achieved high-quality results, they are generally limited to synthesizing novel views with only minor deviation from the input camera position. Consequently, these techniques cannot reconstruct a full 360\textdegree{} scene, which is the primary goal of our work.

With the introduction of diffusion models, the more recent approaches have focused on utilizing these powerful models for 3D scene reconstruction. Specifically, several methods~\cite{li2024scenedreamer360, zhou2025dreamscene360, zhou2024holodreamer} propose various ways to generate 3D scenes from input text prompts. These methods first generate entire panorama from text prompt using pretrained text-to-panorama diffusion models (DMs) and then lift it to 3D. Unfortunately, these approaches are fully generative and do not have a mechanism for reconstructing a 3D scene which is also consistent with a single input image.

Several methods~\cite{chung2023luciddreamer, shriram2024realmdreamer, yu2024wonderjourney, ouyang2023text2immersion, hollein2023text2room, zhang2024text2nerf} specifically address the problem of 3D scene reconstruction from a single image. Starting from the input image, these methods typically project it into 3D space, render it from a novel view, and then inpaint the missing regions using a diffusion model. They repeat this process for a series of cameras along a specific path to reconstruct the complete 3D scene. However, a major limitation of these approaches is that, due to the progressive nature of the scene building, they often fail to synthesize coherent 360\textdegree{} scenes, i.e., the start and end of the 360\textdegree{} scenes are contextually different.

In this work, we propose a novel framework, coined \method{}, for generating a coherent 360\textdegree{} 3D scene from a single input image. Departing from the existing methods, which generate the 3D scene one image at a time, we start by producing a coherent 360\textdegree{} panorama from the input image using standard pre-trained inpainting diffusion models. Inspired by MultiDiffusion~\cite{bar2023multidiffusion}, we formulate the problem as an optimization with two loss terms and propose an alternating minimization strategy to optimize the objective, resulting in a coherent and seamless panoramic image.

The next stage of our approach involves estimating the depth of the panoramic image to project pixels into 3D space and reconstruct the 3D scene. While powerful monocular depth estimation methods~\cite{yang2024depth} exist, these techniques are typically optimized for specific resolutions and struggle to handle large panoramic images effectively. To address this problem, we formulate panoramic depth reconstruction as an optimization task, aiming to simultaneously produce a coherent panoramic depth map and a parametric function that aligns the range of monocular depth to the target depth. We propose an alternating minimization approach to efficiently solve this objective, resulting in a coherent and seamless panoramic depth map.

Given the panoramic image and depth, we directly apply the approach of Shih et al.~\shortcite{shih20203d} to construct a layered depth image (LDI) and inpaint the missing regions in each layer. Next, we build a 3D Gaussian splatting (3DGS) representation~\cite{kerbl20233d} by initializing a set of Gaussians through the projection of LDI pixels into 3D space. We then optimize the 3DGS representation to sharpen details and obtain the final scene. We demonstrate that \method{} can reconstruct consistent 360° 3D scenes from single input images that outperform existing methods. In summary, our work makes the following contributions:

\begin{itemize}
\item We propose a novel framework for synthesizing a coherent 3D panoramic scene from a single image.

\item We formulate the problem of single-image panorama generation using an inpainting diffusion model as an optimization task and solve it using an alternating minimization strategy.

\item We frame the task of obtaining panoramic depth from existing monocular depth estimation methods as an optimization problem and propose an alternating minimization method to solve it.

\end{itemize}

\section{Related Work}

\subsection{Panorama Generation}

Diffusion models (DMs) have shown promising results across various generative tasks. In particular, several approaches~\cite{bar2023multidiffusion, lee2023syncdiffusion, deng2023mv, ye2024diffpano, li2023panogen, frolov2024spotdiffusion, wang2024customizing, zhang2023diffcollage} have proposed leveraging pretrained DMs to synthesize panoramic images. For example, DiffCollage~\cite{zhang2023diffcollage} reconstructs complex factor graphs and aggregates intermediate output from DMs defined by nodes to generate a panorama. PanoGen~\cite{li2023panogen} utilizes latent diffusion models combined with recursive outpainting to create indoor panoramic images. MultiDiffusion~\cite{bar2023multidiffusion} frames the problem of panoramic image generation from pretrained DMs as an optimization process to produce globally consistent images. SynDiffusion~\cite{lee2023syncdiffusion} builds on this idea and incorporates the LPIPS score~\cite{zhang2018unreasonable} between neighboring denoised images into the optimization process. StitchDiffusion~\cite{wang2024customizing} further proposes averaging the overlapping denoising predictions and fine-tuning a low-rank adaptation (LoRA) module~\cite{hu2021lora}. To improve the efficiency of the generation process, SpotDiffusion~\cite{frolov2024spotdiffusion} shifts non-overlapping denoising windows over time to synthesize a coherent panorama efficiently. All of these methods generate panoramas from a text prompt and cannot incorporate an input image into the generation process.

In contrast to these approaches, PanoDiffusion~\cite{wu2023panodiffusion} is designed to generate panoramas from a masked input image. Similarly, MVDiffusion~\cite{deng2023mv} can produce a panorama from a single image by stitching multiple images using pixel-wise correspondences and attention modules. However, both of these approaches require training and struggle to generalize to diverse scenarios.

\subsection{View synthesis from a single input image}

Numerous methods have been proposed to synthesize scenes from a single input image. One category of these methods\edit{~\cite{niklaus20193d, shih20203d, kopf2020one, jampani2021slide, pu2023sinmpi}} addresses this problem in a modular manner, decomposing the synthesis process into several independent components. For example, Shih et al.~\shortcite{shih20203d} estimate a layered depth image (LDI) representation to reconstruct novel views. Niklaus et al.\shortcite{niklaus20193d} use the estimated depth to map the input image to a point cloud and train a network to fill in the missing areas.

The second group of methods~\cite{zhou2016view, srinivasan2017learning, li2020synthesizing, tucker2020single} synthesizes scenes from a single input image in an end-to-end manner. Among these approaches, Zhou et al.\shortcite{zhou2016view} propose synthesizing scenes by first estimating optical flow and then warping the input image to novel views. Srinivasan et al.\shortcite{srinivasan2017learning} use two sequential convolutional neural networks to estimate disparity and refine the warped images. Several approaches propose synthesizing intermediate scene representations to achieve view synthesis. For example, SynSin~\cite{wiles2020synsin} estimates a point cloud of a scene, and several methods~\cite{li2020synthesizing, tucker2020single} synthesize light fields using the estimated multi-plane image (MPI) representation. PixelNeRF~\cite{yu2021pixelnerf} trains a NeRF prior and can synthesize NeRF from a single input image without performing test-time optimization. Additionally, several approaches~\cite{paliwal2023reshader, bello2024novel} focus on improving the view-dependent effects for single-view view synthesis. However, all of these methods are designed only for view synthesis within a narrow angle or restricted camera movement and cannot be generalized to the entire 360\textdegree{} scene.

\subsection{3D Scene Generation}

Reconstructing an entire 3D scene is a challenging problem, as it requires maintaining both content and depth consistency across a wide range of camera trajectories. Many approaches have been proposed to achieve 3D scene generation, typically leveraging pretrained, powerful 2D diffusion priors, such as latent diffusion models (LDMs), to synthesize 3D scenes by optimizing different 3D representations, such as NeRF and 3DGS. These approaches can be categorized into two groups based on the input condition.

The first group of methods\edit{~\cite{ouyang2023text2immersion, shriram2024realmdreamer, zhang2024text2nerf, engstler2024invisible, yu2024wonderjourney, yu2024wonderworld, chung2023luciddreamer, liang2025wonderland, ni2025wonderturbo}} generates 3D scenes from text or images in a progressive manner. Starting from a single image, either provided by the user or generated from a text prompt, these methods typically perform progressive inpainting, monocular depth estimation, and 3D optimization for novel views in the 3D scene. These approaches differ in their 3D representation, image inpainting, and depth refinement strategies. However, since the 3D scene is generated through progressive inpainting of single inputs, these methods struggle to preserve coherency, making it difficult to synthesize consistent 360\textdegree{} scenes.

The second group of methods~\cite{li2024scenedreamer360, zhou2025dreamscene360, zhou2024holodreamer} generates 360° 3D scenes in a two-step process. They synthesize coherent panoramas by leveraging pretrained text-to-panorama DMs, which are then lifted to 3D using different inpainting and depth estimation strategies. Although these approaches are capable of generating consistent 3D scenes from inputs, they are text-conditioned only and do not have any mechanism to reconstruct a scene consistent with a single input image. In comparison, our method, \method{}, not only generates coherent 3D scenes but also allows users to condition the generation on any single input image.

\section{Preliminaries}
\label{sec:bg}
% \subsection{MultiDiffusion}

In this section, we describe MultiDiffusion~\cite{bar2023multidiffusion}, an approach that leverages a pre-trained diffusion model, without any fine-tuning, to produce results in various image or condition spaces. For example, this technique can generate outputs at resolutions different from the base model’s native resolution (e.g., panoramas) or synthesize images using region-based text prompts. Here, we focus our discussion on the former example, as it is most relevant to our approach.

MultiDiffusion uses a pre-trained diffusion model, $\Phi$, which operates on images of size $H\times W$ as the base model. Starting with an image $I_T$ initialized with Gaussian noise and conditioned on a text prompt $p$, the base model iteratively denoises $I_T$, producing a sequence of intermediate images $I_{T-1}, \cdots, I_1$ and ultimately generating a clean image $I_0$ as follows:

% \vspace{-0.1in}
\begin{equation}
    I_{t-1} = \Phi(I_t \vert p).
\end{equation}
% \vspace{-0.1in}

The goal of MultiDiffusion is to leverage this base model to generate an image $J_0$ at a larger resolution $H^\prime \times W^\prime$. The MultiDiffusion process begins with a noisy high-resolution image, $J_T$, and produces a clean image $J_0$ through a sequence of gradually denoised images $J_{T-1}, \cdots, J_0$. Given the optimal high-resolution image at the current step, $J^*_t$, the key idea of MultiDiffusion is to ensure that the output of the base diffusion model $\Phi(F_i(J^*_t) \vert p)$ is as close as possible to the high-resolution image at the next step $F_i(J_{t-1})$, locally. Note that $F_i$ is an operator that maps the high-resolution image space to the base model's space (via cropping, in this case). Enforcing this similarity in the $L_2$ sense, we arrive at the following objective:

% \vspace{-0.1in}
\begin{equation} 
\label{eq:multiDM_pano}
   J^*_{t-1} = \arg\min_{J} \sum_{i=1}^n \left\Vert W_i \odot \left[ F_i(J) - \Phi(F_i(J^*_t) \vert p)\right] \right\Vert^2,
\end{equation}
% \vspace{-0.1in}
 
\noindent where $W_i$ is a weight map ($W_i = \mathbf{1}$ in this case), $n$ refers to the total number of crops, and $\odot$ denotes the element-wise product.

Since this objective is quadratic, the solution can be easily obtained in closed form as follows:

% \vspace{-0.15in}
\begin{equation}
\label{eq:multiDM_solution}
    J^*_{t-1} = \sum_{i=1}^n\frac{F_i^{-1}(W_i)}{\sum_{j=1}^n F_j^{-1}(W_j)} \odot F_i^{-1}(\Phi(F_i(J^*_t) \vert p)),
\end{equation}
% \vspace{-0.05in}

\noindent where $F_i^{-1}$ is the inverse of the cropping operator, which places the content into the appropriate location in the high-resolution image. At a high level, this solution aggregates (adds) the outputs of the base diffusion model for overlapping crops and normalizes the resulting image by the total number of crops at each pixel.

Starting from the noisy high-resolution image $J^*_T = J_T$, MultiDiffusion uses this process to obtain the optimal intermediate high-resolution images $J^*_t$, resulting in the final clean high-resolution image $J^*_0$.

\section{Algorithm}

Given a single input image $I$, our goal is to reconstruct a coherent 360 scene using a 3D Gaussian representation~\cite{kerbl20233d}. Unlike existing methods that produce the 3D scenes through progressive projection and inpainting, we begin by generating a coherent 360\textdegree{} panorama from the input image (Sec.~\ref{ssec:panimg}). We then estimate a coherent and consistent depth from the generated panorama (Sec.~\ref{ssec:pandepth}). Finally, we inpaint the occluded regions using layered depth image (LDI) inpainting and use the inpainted layers to reconstruct a 3DGS representation (Sec.~\ref{ssec:inpaint}).

\subsection{Single-Image Panorama Generation}
\label{ssec:panimg}

\begin{figure}
    \centering
    \includegraphics[width=\linewidth]{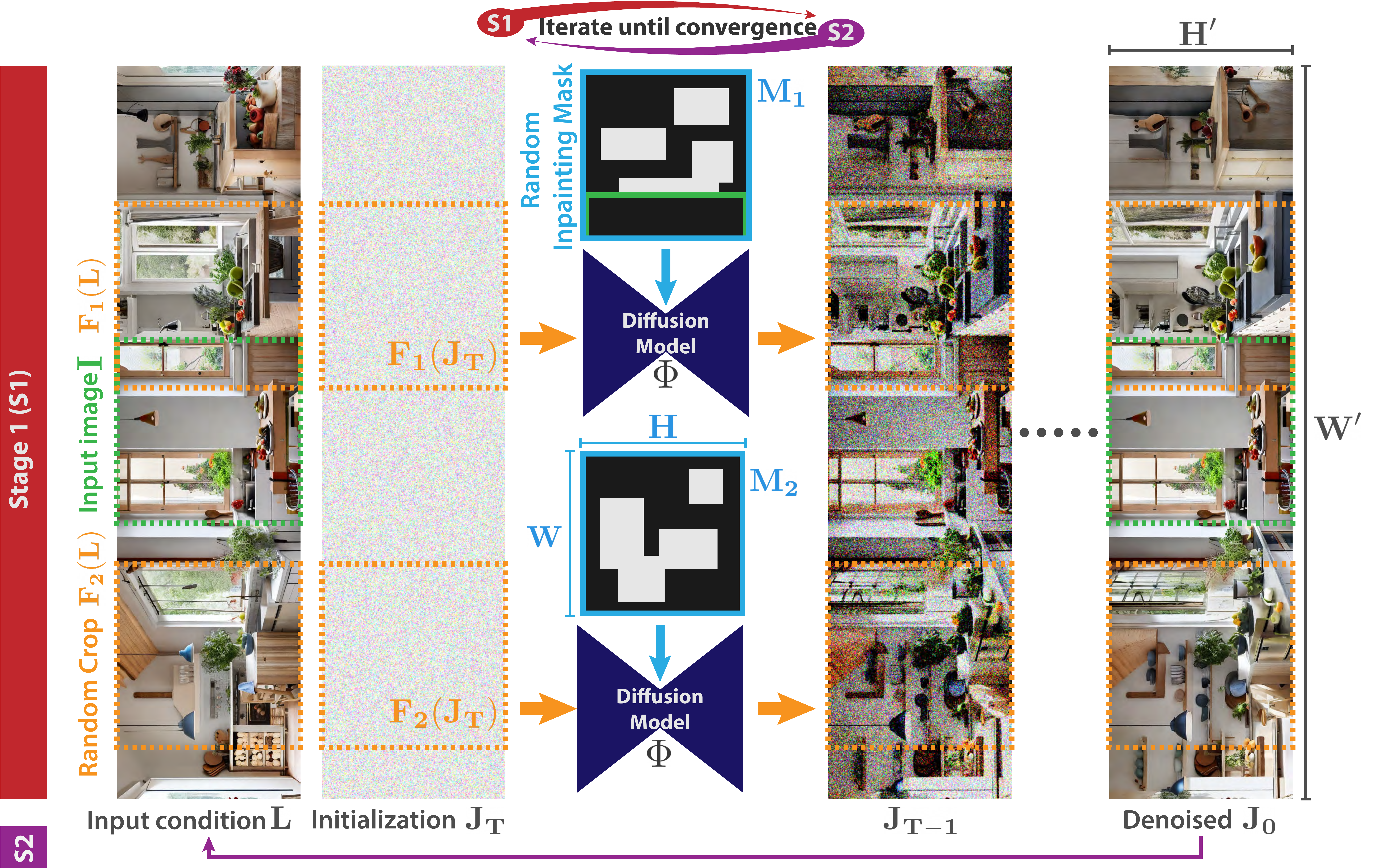}
    \vspace{-0.25in}
    \caption{We provide an overview of our proposed \PanoMethod{} process, which consists of two stages. In the first stage, we fix the input condition $L$ and apply the diffusion model to overlapping crops of the image at the current time step. The outputs are then aggregated to produce the image at the next time step. This process is repeated until the fully denoised image $J_0$ is obtained. In the next stage, we replace the current input condition with $J_0$. These two stages are repeated until convergence.}
    \label{fig:pano_overview}
    \Description{}
    \vspace{-0.25in}
\end{figure}

\begin{figure}
    \centering
    \includegraphics[width=\linewidth]{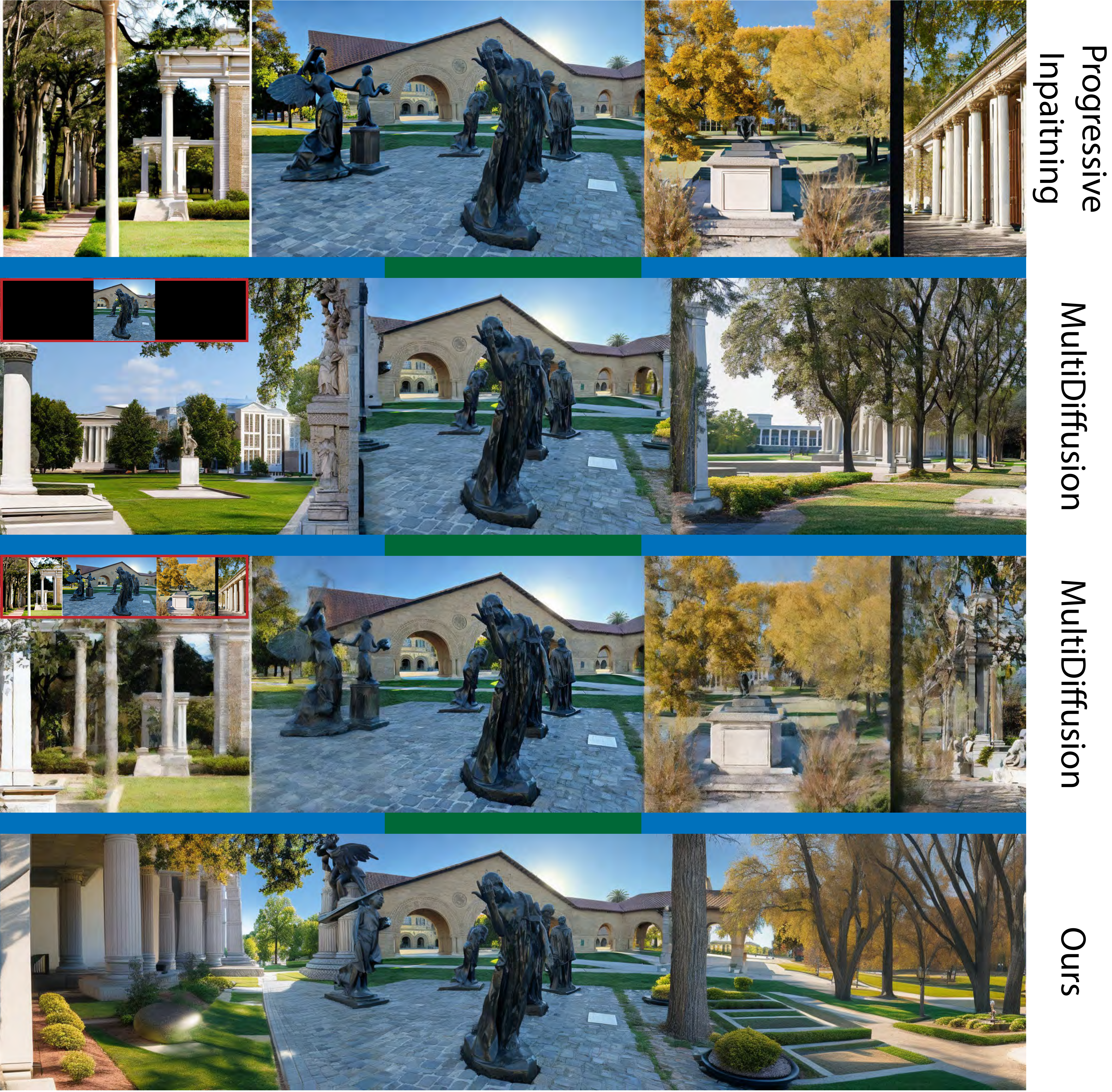}
    \vspace{-0.25in}
    \caption{We compare the results of our \PanoMethod{} process against MultiDiffusion and progressive inpainting. The green bar shows the location of the input image. We show MultiDifussion results with two different input conditions (shown on the top left): black canvas with input image (second row), and progressive inpainting result (third row). Our method produces coherent results, while the alternative approaches produce images with seams and inconsistencies.}
    \label{fig:pano_approach}
    \Description{}
    \vspace{-0.2in}
\end{figure}

We begin by discussing our method for generating a larger image from a single input image, then explain the specific details for panorama generation in Sec.~\ref{ssec:panimgdet}. The overview of our approach is provided in Fig.~\ref{fig:pano_overview}. Given an input image $I$ placed on a larger canvas $L$ of size $H' \times W'$, our goal is to fill in missing areas in $L$ using an inpainting diffusion model $\Phi$, which operates on fixed lower-resolution images of size $H \times W$. In addition to a text prompt $p$, this model takes a mask $M$ denoting the missing regions and a masked image $(1 - M) \odot X$ as inputs. It progressively denoises a Gaussian noise image $I_T$ to obtain a clean image $I_0$ containing the hallucinated details, with each step following $I_{t-1} = \Phi(I_t | p, M, (1-M) \odot X)$. A straightforward approach is to use this model to gradually outpaint the high-resolution image, starting from the regions covered by the input. However, this approach often results in noticeable contextual inconsistencies and seams, as shown in Fig.~\ref{fig:pano_approach} (Progressive Inpainting).

Inspired by MultiDiffusion, we address this issue by formulating the problem as an optimization. MultiDiffusion can be adapted to this problem in a straightforward manner by replacing the base diffusion model with an inpainting model and reformulating the objective in Eq.~\ref{eq:multiDM_pano} as follows:

% \vspace{-0.1in}
\begin{equation}
    \mathcal{L}(J_{t-1}\vert J^*_t) = \hspace{-0.05in} \sum_{i=1}^n \Vert M_{i} \odot \left[ F_i(J_{t-1}) - \Phi(F_i(J^*_t)\vert \mathcal{C}_{i})\right] \Vert^2\hspace{-0.05in},
\end{equation}
% \vspace{-0.15in}

\noindent where

% \vspace{-0.15in}
\begin{equation}
\label{eq:condition}
    \mathcal{C}_{i} = \{p, M_{i}, M_{i} \odot F_i(L) \}.
\end{equation}
% \vspace{-0.1in}

\noindent Here, $M_i$ is a random inpainting mask for the $i^\text{th}$ crop, and $L$ is the high-resolution condition image. This objective ensures that the output of the inpainting diffusion model, $\Phi(F_i(J^*_t)\vert \mathcal{C}_{i})$, is as close as possible to the corresponding crop of the high-resolution image at the next step $F_i(J_{t-1})$ in the masked areas $M_i$. This objective can be minimized similarly to Eq.~\ref{eq:multiDM_solution} as follows:

% \vspace{-0.0in}
\begin{equation}
\label{eq:multiCDM_solution}
    J^*_{t-1} = \sum_{i=1}^n\frac{F_i^{-1}(M_i)}{\sum_{j=1}^n F_j^{-1}(M_j)} \odot F_i^{-1}(\Phi(F_i(J^*_t)\vert \mathcal{C}_{i})),
\end{equation}
% \vspace{-0.0in}

Based on this equation, it is easy to infer that the solution depends on the high-resolution input condition, $L$, of the MultiDiffusion process. As shown in Fig.~\ref{fig:pano_approach}, MultiDiffusion results vary drastically depending on how the input condition is set. In particular, both simple methods for obtaining the input condition, such as placing the input image on a black canvas or using progressive inpainting, produce inconsistent results.

To address this issue, we make a key observation that the ideal input condition is a coherent and consistent high-resolution image. However, obtaining such an image, $J_0$, is the goal of our optimization and is not available beforehand. Therefore, we propose to incorporate this observation as an additional term in our objective as follows:

% \vspace{-0.15in}
\begin{equation} 
\label{eq:multicondDM}
   {\tilde{J}_0 \Compactdots \tilde{J}_{T-1}, \tilde{L} = \argmin_{J_0 \Compactdots J_{T-1}, L} \left[ \sum_{t = T}^{1}\mathcal{L}(J_{t-1}\vert J^*_t) + \Vert L - J_0 \Vert^2 \right]}
\end{equation}
% \vspace{-0.10in}

\noindent where the first term is the adapted MultiDiffusion objective for all time steps, while the second term forces the condition image $L$ to be close to the clean high-resolution image $J_0$. Note that the output of this process, $J_{T-1}, \dots, J_0$, depends on the condition image $L$. As such, $J^*_t$ is the optimal solution at time $t$ given the current condition image $L$, and it differs from the final optimal solution, $\tilde{J}_t$, which is obtained using the optimal condition image $\tilde{L}$. We call this equation the \emph{\PanoMethod{}} objective, as the high-resolution diffusion process in our case is conditional.

Simultaneously solving for all the images in this objective is a difficult task. Therefore, we propose an alternating minimization strategy that solves for $J_{T-1}, \dots, J_0$ and $L$ in the following two stages:

% \vspace{-0.05in}
\paragraph{Stage 1:} Here, we fix $L$ and minimize Eq.~\ref{eq:multicondDM} by finding the optimal $J_{T-1}, \dots, J_0$. Since $J_{T-1}, \dots, J_1$ do not influence the second term (as different steps are assumed to be independent), we can use Eq.~\ref{eq:multiCDM_solution} to obtain their solution in closed form. On the other hand, since $J_0$ appears in both terms, and both terms are quadratic with respect to it, the final solution is a weighted combination of the solution to the first term (Eq.~\ref{eq:multiCDM_solution}) and the second term ($J^*_0 = L$). In practice, however, we found that plausible results can still be obtained even when the second term is ignored. 

To summarize, as shown in Fig.~\ref{fig:pano_overview}, starting from $J_T$, we aggregate the output of the inpainting diffusion model over different overlapping crops to obtain the image at the next time step, resulting in a sequence of optimal $J^*_{T-1}, \dots, J^*_0$ given the current fixed high-resolution input condition $L$.

% \vspace{-0.05in}
\paragraph{Stage 2:} During this stage, we fix $J_{T-1}, \dots, J_0$ and find the optimal $L$ that minimizes Eq.~\ref{eq:multicondDM}. $L$ influences both the first term, as the diffusion model is conditioned on it (see Eq.~\ref{eq:condition}), and the second term. Obtaining the optimal solution to each term independently is straightforward. The optimal solution to the first term is $L^* = L$, since if $L$ was used to produce the current $J_{T-1}, \dots, J_0$, it is likely the best option for reproducing the same results. Moreover, since the second term is quadratic, the solution is simply $L^* = J_0$. Although obtaining the solution to each term is straightforward, computing the optimal solution considering both terms is difficult. However, assuming that $L$ and $J_0$ are close to each other---i.e., \PanoMethod{} does not diverge significantly from the condition image in one pass---it is reasonable to assume that $L^* = J_0$ is close to the optimal solution for both terms.

We perform the optimization by first initializing $J_T$ with Gaussian noise and $L$ by placing the input image on a black canvas. We then alternate between stages 1 and 2 iteratively until convergence. At the end of this process, we can use either $\tilde{J}_0$ or $\tilde{L}$ as the final result. Fig.~\ref{fig:pano_approach} compares \PanoMethod{} with MultiDiffusion and progressive inpainting.

\begin{figure}
    \centering
    \includegraphics[width=\linewidth]{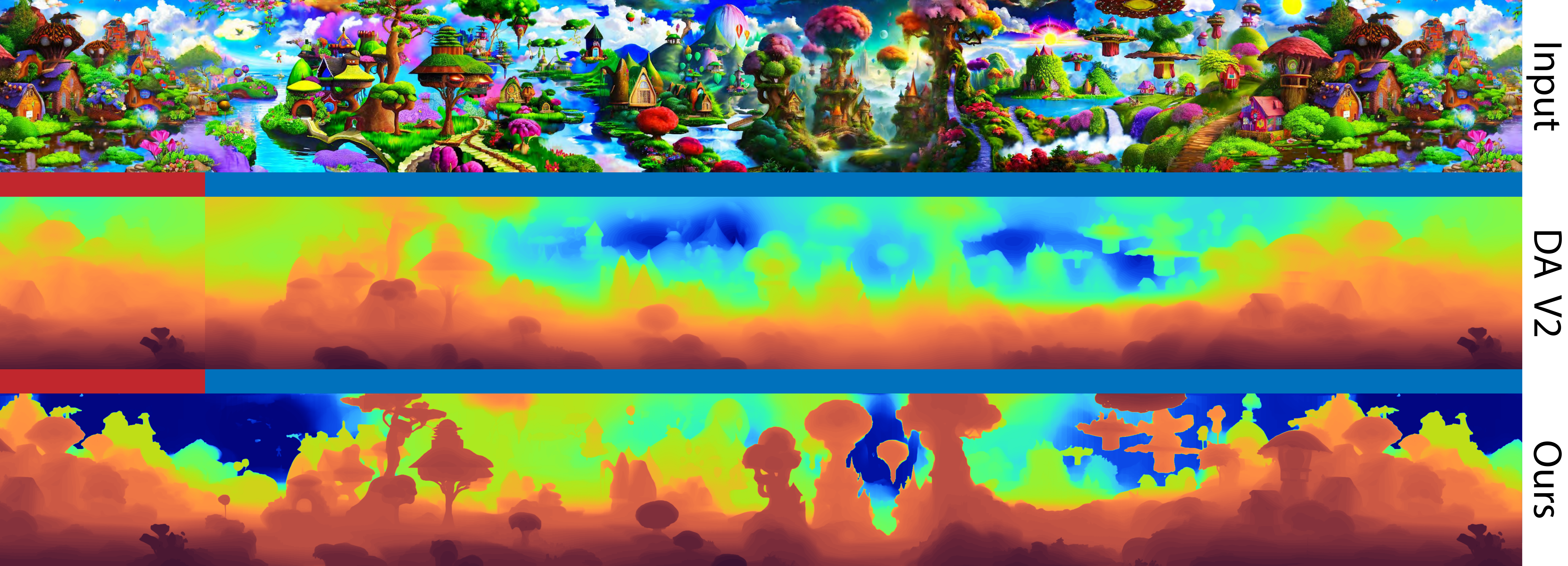}
    \vspace{-0.25in}
    \caption{We compare the result of our method, \DepthMethod{}, against applying Depth Anything V2 (DA V2)~\cite{yang2024depthv2} on the full image. The results obtained by DA V2 lacks details and is geometrically inconsistent. Our approach, on the other hand, produces highly detailed and consistent depth maps.}
    \label{fig:DA_highres_comp}
    \Description{}
    \vspace{-0.1in}
\end{figure}

\begin{figure}
    \centering
    \includegraphics[width=\linewidth]{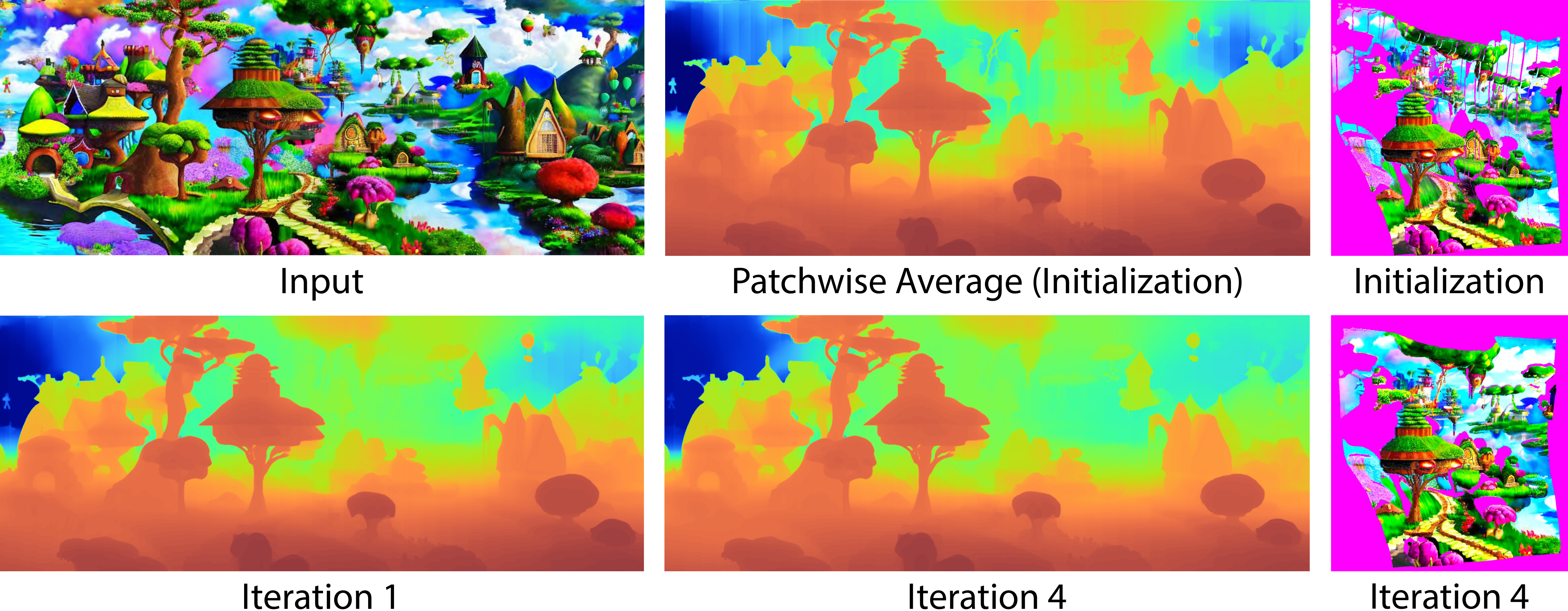}
    \vspace{-0.2in}
    \caption{Averaging the patch depth estimates leads to banding artifacts since the depth maps are relative and not consistent. On the top right, we show that projecting the image into 3D using such a depth map results in clear banding artifacts. Since we initialize $G_{\theta_i}$ with the identity line, the patchwise average serves as our initial depth estimate during the optimization of Eq.~\ref{eq:pano_depth}. We also show our results after one and four iterations of optimization. After only four iterations, the seams disappear. As seen on the bottom right, the banding artifacts also disappear from the projected image.}
    \label{fig:depth_iterations}
    \Description{}
    \vspace{-0.1in}
\end{figure}

\begin{figure}
    \centering
    \includegraphics[width=\linewidth]{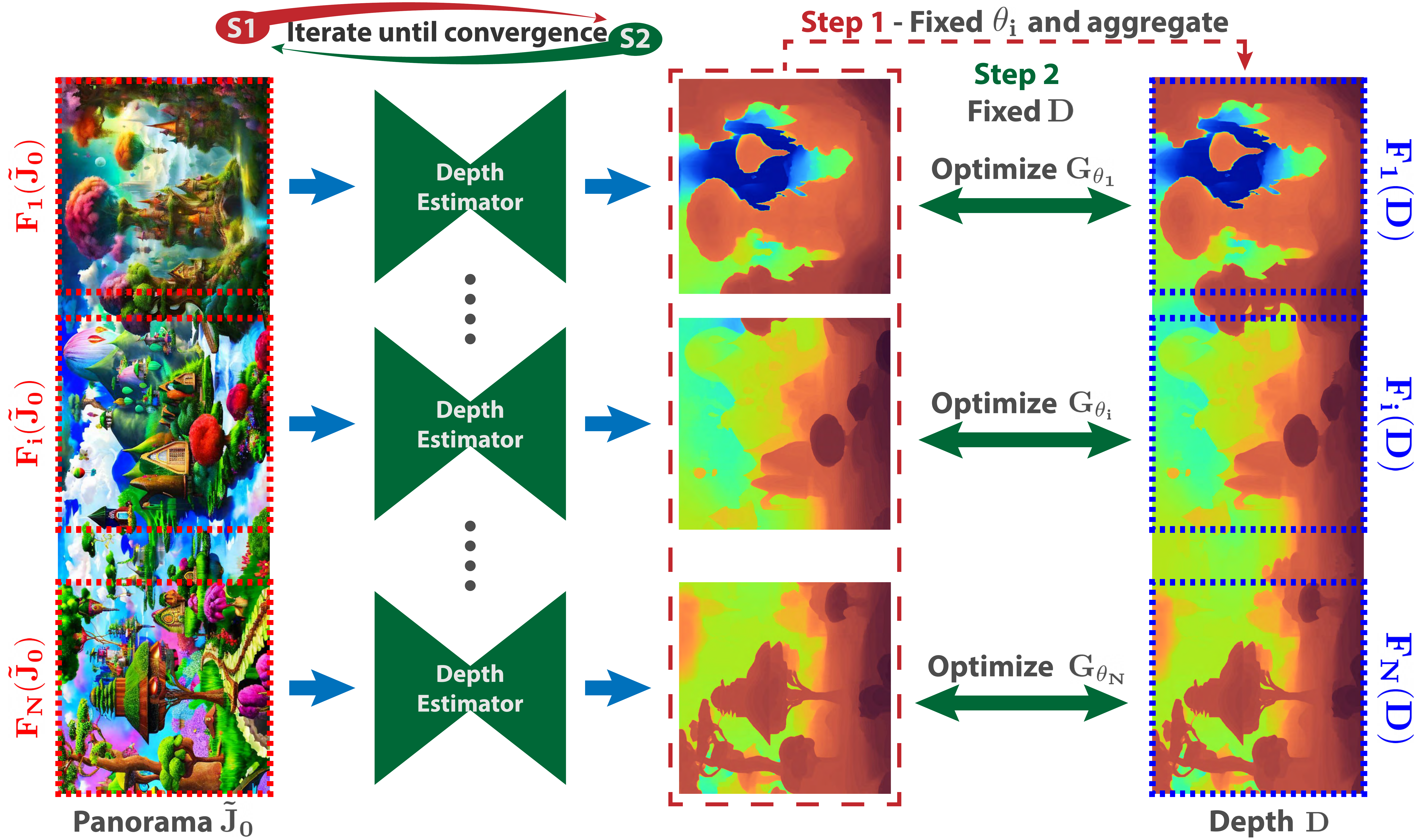}
    \vspace{-0.25in}
    \caption{We show the overview of \DepthMethod{}. We first apply an existing depth estimator to the overlapping patches of the input image to obtain a set of patch depth estimates. We then perform optimization in two stages. In the first stage, the depth patches are adjusted using a piecewise linear function $G_{\theta_i}$, and the adjusted patches are then aggregated to obtain the panoramic depth. In the second stage, we optimize the parameters $\theta_i$ of the parametric functions to match the adjusted patch depth estimates with the corresponding regions in the panoramic depth. These two steps are repeated until convergence.}
    \label{fig:depth_overview}
    \Description{}
    \vspace{-0.2in}
\end{figure}

\begin{figure*}
    \centering
    \includegraphics[width=\linewidth]{figures/3dgs_overview.pdf}
    \vspace{-0.25in}
    \caption{We present an overview of our inpainting and 3DGS optimization process. Given a cylindrical panorama and its corresponding depth, we first convert them to the LDI representation. We then inpaint both the image and depth layers. Note that while all images and depth maps are cylindrical, we show only a small crop for clarity. Next, we initialize the Gaussians by assigning a single Gaussian to each pixel and projecting them into 3D space. Finally, we perform 3DGS optimization to obtain the 3D representation.}
    \label{fig:3dgs_overview}
    \Description{}
    \vspace{-0.1in}
\end{figure*}

\begin{table*}[h]
    \centering
    \caption{Numerical comparison of \PanoMethod{} for single image wide-image generation against other relevant methods. CLIP-IQA+ and Q-Align measure the quality, A-CLIP and A-Align asses the aesthetic, and C-CLIP and C-Style evaluate the consistency of the results.}
        % \vspace{-0.1in}
        
    \resizebox{0.7\linewidth}{!}{
    \begin{tabular}{l|cccccc}
        \hline
        Method & Q-IQA $\uparrow$ & Q-Align $\uparrow$ & A-CLIP $\uparrow$ & A-Align $\uparrow$ & C-CLIP $\uparrow$ & C-Style $\downarrow$ \\
        \hline
        Progressive & 0.520 & 4.164 & 5.779 & 3.314 & 0.862 & 0.019  \\
        L-MAGIC~\cite{cai2024magic} & \textbf{0.550} & 4.331 & 5.865 & 3.426 & 0.842 & 0.069 \\
        MultiDiffusion~\cite{bar2023multidiffusion} & 0.523 & 4.390 & 5.953 & 3.516 & 0.869 & 0.030 \\
        SyncDiffusion~\cite{lee2023syncdiffusion} & 0.535 & 4.290 & 5.893 & 3.429 & 0.864 & 0.016 \\
        \PanoMethod{} (ours) & 0.530 & \textbf{4.481} & \textbf{5.992} & \textbf{3.696} & \textbf{0.881} & \textbf{0.011} \\
        \hline
    \end{tabular}}
    \label{table:pano}
\end{table*}

\subsubsection{Panorama Generation Details}
\label{ssec:panimgdet}

We slightly modify the \PanoMethod{} process to adapt it for generating panoramas from a single image. Our goal is to produce a cylindrical panorama, so in this case, \PanoMethod{} operates in the cylindrical domain, and the sequence $J_T, \dots, J_0$ is defined within this domain. Since the base inpainting diffusion model operates on perspective images, $F_i$ performs both cropping and projection from the cylindrical to the perspective domain. Similarly, $F_i^{-1}$ projects the pixels from the perspective image back to the cylindrical image, placing them in the appropriate locations. 

We experimented with bilinear interpolation during the projection; however, interpolation smoothed out the noise, which negatively affected the performance of the diffusion model. Therefore, we instead use nearest-neighbor interpolation for both $F_i$ and $F_i^{-1}$. Additionally, we use an FOV of 45\textdegree{} for the perspective camera and carefully set the resolution of the cylindrical image to ensure a near one-to-one mapping between the pixels of the cylindrical and perspective images to preserve the noise pattern during projections.

This process allows us to produce a contextually coherent and seamless 360\textdegree{} cylindrical panorama, which we use to reconstruct the 3D scene. In our experiments, we apply 20 iterations of \PanoMethod{} (Stage 1 + Stage 2) to obtain the final cylindrical panorama.

\subsection{Panorama Depth Estimation}
\label{ssec:pandepth}

Given the panoramic image $\tilde{J}_0$, our goal is to estimate its depth $D$. In recent years, several powerful monocular depth estimation methods~\cite{yang2024depthv2, yang2024depth} have been introduced. These approaches can estimate highly detailed relative depth but typically perform best at a specific image size. Beyond this optimal resolution, they often produce results that lack detail and geometric consistency. Consequently, applying these methods directly to panorama depth estimation leads to poor results, as shown in Fig.~\ref{fig:DA_highres_comp}.

We address this problem by obtaining $D$ through a combination of estimated depth maps on patches using an existing technique, $\Psi$, i.e., $\Psi(F_i(\tilde{J}_0))$. However, na\"ively combining the patches (e.g., through averaging) leads to unsatisfactory results (see Fig.~\ref{fig:depth_iterations}), as the patch depth estimates are relative and can be inconsistent. To overcome this, we pose the problem of obtaining panoramic depth from patch depth estimates as an optimization task. Our key insight is that the panoramic depth $D$ should be close to the estimated depth \emph{after} it has been globally aligned through a parametric function. This can be formally written as:

% \vspace{-0.1in}
\begin{equation}
\label{eq:pano_depth}
    {D^*, \bm{\theta}^* = \argmin_{D, \bm{\theta}}\sum_{i=1}^n\Vert F_i(D) - G_{\theta_i}(\Psi(F_i(\tilde{J}_0))) \Vert^2,}
\end{equation}
% \vspace{-0.05in}

\noindent where $G_{\theta_i}$ is the parametric function, and $\bm{\theta} = \{ \theta_1, \dots, \theta_n\}$ represents the set of parameters for different patches. In our implementation, we use a piecewise linear function, where each parameter consists of a series of scale and shift values.

Solving for both $D$ and $\bm{\theta}$ simultaneously is challenging. Therefore, we propose performing this optimization through alternating minimization, consisting of two stages (see Fig.~\ref{fig:depth_overview}). In the first stage, we fix $\bm{\theta}$ and find the optimal $D$. Since the objective is quadratic, the solution can be obtained in closed form, similar to Eq.~\ref{eq:multiDM_solution}. The only difference is that $\Phi$, the diffusion model, is replaced with $\Psi$, and $W_i = \bm{1}$. In the second stage, we fix $D$ and find the optimal $\bm{\theta}$. This is a least-squares regression problem, which we solve using standard packages.

Starting with all $G_{\theta_i}$ as the identity line (i.e., a linear function with a slope of 1), we alternate between the first and second stages iteratively until convergence (four iterations in our implementation). Once converged, we obtain a coherent and consistent panoramic depth, as shown in Figs.~\ref{fig:DA_highres_comp}~and~\ref{fig:depth_iterations}.

\subsection{Inpainting and 3DGS Optimization}
\label{ssec:inpaint}

We now discuss our process for reconstructing the 3D scene using our generated panorama and the corresponding depth map (see the overview in Fig.~\ref{fig:3dgs_overview}). While the estimated depth can be used to project the cylindrical image into 3D space, when the scene is viewed from any position other than the panorama's center of projection, occluded regions become visible. To address this, we utilize the layered depth image (LDI) inpainting approach by Shih et al.~\shortcite{shih20203d}, which performs effective depth-aware texture inpainting while also providing the corresponding depth. We use a four-layer LDI representation (foreground, background, and two intermediate layers) based on agglomerative clustering by disparity. 

We then use these cylindrical layered images and depth maps to initialize a set of 3D Gaussians. Specifically, we assign a Gaussian to each pixel of the image at each layer and project them into 3D according to the corresponding depth. The color of each Gaussian is initialized based on the pixel color (without spherical harmonics); we initialize the rotation matrix with identity, assign the scale following Paliwal et al.~\shortcite{paliwal2025coherentgs}, and set the opacity to 0.5. During this process, we keep track of which Gaussians correspond to which layer, as this information is required for optimization.

Next, we perform 3DGS optimization for 1000 iterations. To do this, we set up 240 evenly rotated cameras from the center of projection and project the layered images and depth maps to these cameras. During the optimization, we randomly sample one of these cameras and optimize the Gaussians according to their corresponding layer. Additionally, we composite all the four layers and use the composited image as a reference to optimize all the Gaussians. \edit{Note that per layer and composite losses are optimized simultaneously.} We use the original 3DGS reconstruction loss along with an $L_2$ loss between the rendered and layered depth maps. In addition, to be able to produce consistent results from novel views, we use the depth-based novel view loss, proposed by Zhu et al.~\shortcite{zhu2024fsgs}. \edit{Furthermore, pruning and densification are performed following 3DGS.} Once the optimized 3DGS representation is obtained, we can synthesize novel views of the scene and produce coherent and seamless results.

\begin{figure*}
    \centering
    \includegraphics[width=\linewidth]{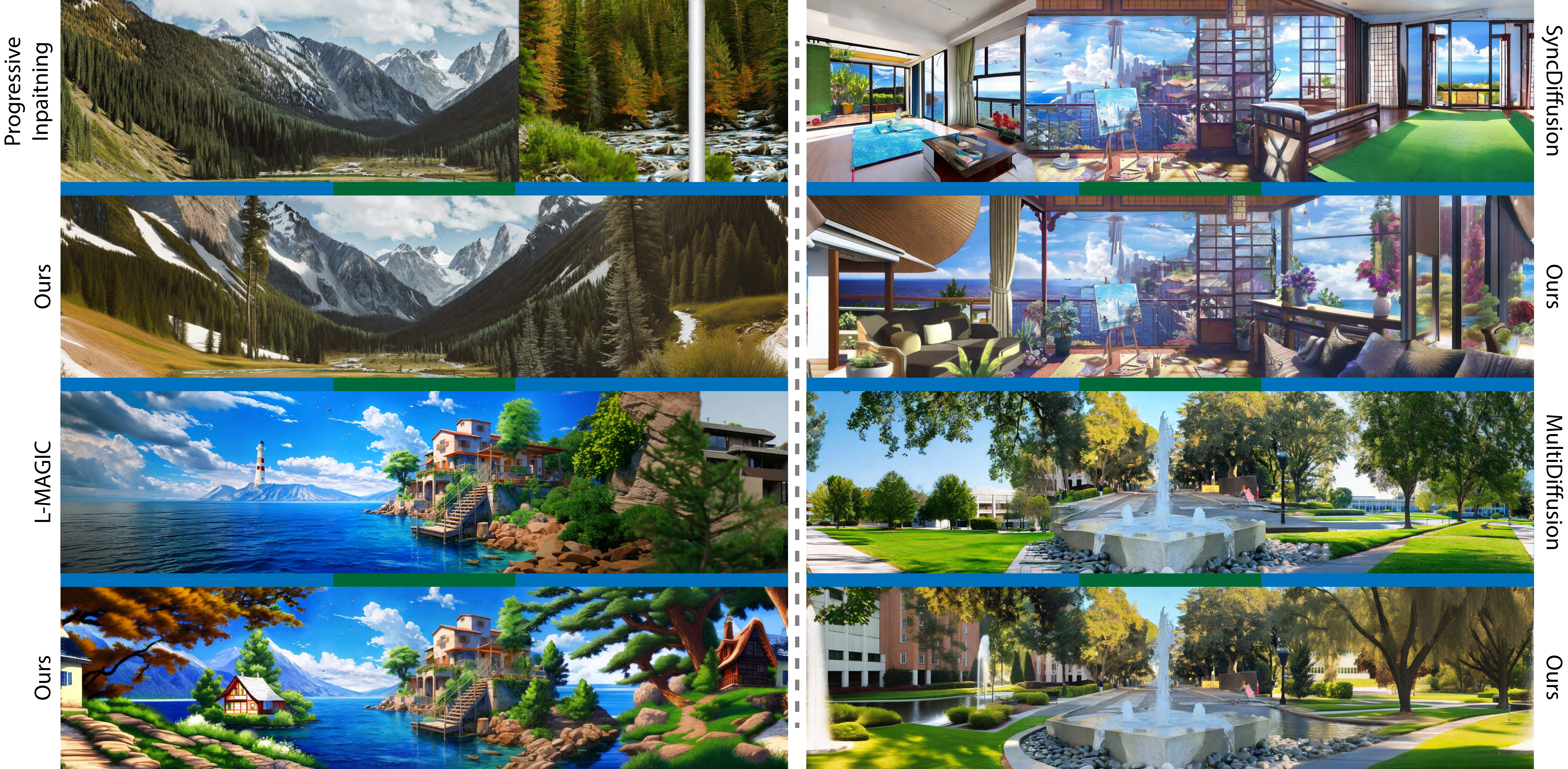}
    \caption{We compare the wide-images generated by \PanoMethod{} with those from other methods. Other approaches often result in sharp discontinuities and contextual inconsistencies. For instance, in the top example, the MultiDiffusion result shows a mismatch between the generated sky and the input sky.}
    \label{fig:pano-comparison}
    \Description{}
    % \vspace{-0.6cm}
\end{figure*}

\section{Results}

In this section, \edit{we evaluate both the generated texture and depth of our proposed algorithms. Specifically,} we compare our approach against state-of-the-art wide-image generation and 3D scene generation methods, both visually and numerically. For evaluation, we compile a test set of 28 real and synthetic scenes sourced from LucidDreamer~\cite{chung2023luciddreamer} and WonderJourney~\cite{yu2024wonderjourney}. For numerical evaluation, we employ several metrics to evaluate different aspects of the results: (1) Quality - we assess the quality of results using CLIP-IQA+~\cite{wang2023exploring} and Q-Align~\cite{wu2023q} scores. CLIP-IQA+ and Q-Align are built upon contrastive language-image pre-training (CLIP)~\cite{radford2021learning} and large multi-modality models (LMMs) for image quality assessment, respectively. (2) Aesthetic - we use the CLIP aesthetic score (A-CLIP)  and A-Align~\cite{wang2023exploring} to measure results aesthetic. (3) Consistency - we compute the similarity (C-CLIP) and style loss~\cite{gatys2016image} (C-Style) of the CLIP embeddings of random pairs of non-overlapping patches in the results.

\begin{table*}[h]
    \label{tab:3d}
    \centering
    
    \caption{Numerical comparisons of our approach against the state-of-the-art methods on novel view synthesis. The evaluation metrics are the same as the ones in Table~\ref{table:pano}.}
    \vspace{-0.1in}
    \resizebox{0.7\linewidth}{!}{
    \begin{tabular}{lcccccc}
        \hline
        Method & Q-IQA $\uparrow$ & Q-Align $\uparrow$ & A-CLIP $\uparrow$ & A-Align $\uparrow$ & C-CLIP $\uparrow$ & C-Style $\downarrow$ \\
        \hline
        LucidDreamer~\cite{chung2023luciddreamer} & 0.495 & 2.911 & 5.253 & 2.705 & 0.848 & 0.058  \\
        WonderJourney~\cite{yu2024wonderjourney} & \textbf{0.504} & \textbf{3.506} & 5.368 & \textbf{2.834} & 0.820 & 0.058 \\
        \method{} (ours) & 0.443 & 3.305 & \textbf{5.673} & 2.772 & \textbf{0.869} & \textbf{0.025} \\
        \hline
    
    \end{tabular}}
    \label{table:3d}
        % \vspace{-0.1in}
\end{table*}

\begin{figure*}
    \centering
    \includegraphics[width=\linewidth]{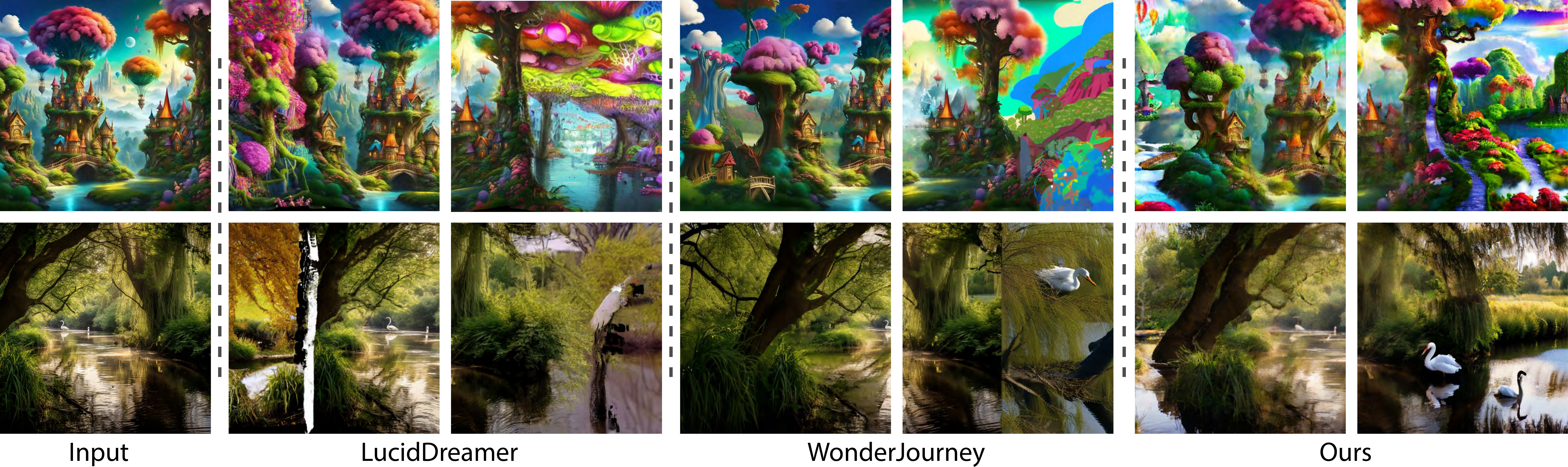}
    \vspace{-0.25in}
    \caption{We compare renderings of \method{} with LucidDreamer~\cite{chung2023luciddreamer} and WonderJourney~\cite{yu2024wonderjourney}. For each methods, we render 3D scene from two novel views. LucidDreamer and WonderJourney produces results with seams and inconsistencies. In comparison, \method{} is capable of generating coherent renderings from novel views. For more visual results and video comparison, please refer to our supplementary materials. }
    \label{fig:3d-comparison}
    \Description{}
    % \vspace{-0.1in}
\end{figure*}

\begin{table}[h]
    \centering
    \caption{\edit{Comparison of \DepthMethod{} with Moge's~\cite{wang2025moge} patch-wise depth estimation + blending approach. Rows $2-3$ correspond to \DepthMethod{} combined with different depth estimators.}}
    \vspace{-0.1in}
    \resizebox{0.9\linewidth}{!}{
    \begin{tabular}{lcc|ccc}
        \hline
        Method & AbsRel $\downarrow$ & RMSE $\downarrow$ & $\delta_{1} \uparrow$ & $\delta_{2} \uparrow$ & $\delta_{3} \uparrow$ \\
        \hline
        MoGe & 0.21 & \textbf{0.56} & 61.66 & 90.63 & 98.32 \\
        DA V2 + ours & 0.31 & 0.91 & 47.05 & 77.39 & 92.17 \\
        MoGe + ours & \textbf{0.19} & 0.67 & \textbf{66.47} & \textbf{92.66} & \textbf{98.71} \\
        \hline
    \end{tabular}
    \label{tab:depth}
    }
        \vspace{-0.1in}
\end{table}

\begin{table*}[h]
    \centering
    
    \caption{\edit{Numerical ablations to evaluate the effectiveness of different components of our novel view synthesis pipeline. The evaluation metrics are the same as the ones in Table~\ref{table:pano}.}}
    \vspace{-0.1in}
    
    \resizebox{0.7\linewidth}{!}{
    \begin{tabular}{lcccccc}
    
        \hline
        Method & Q-IQA $\uparrow$ & Q-Align $\uparrow$ & A-CLIP $\uparrow$ & A-Align $\uparrow$ & C-CLIP $\uparrow$ & C-Style $\downarrow$ \\
        \hline
        
        LucidDreamer~\cite{chung2023luciddreamer} & \textbf{0.495} & 2.911 & 5.253 & 2.705 & 0.848 & 0.058  \\

        LuciDreamer w/ our panorama & 0.469 & 3.012 & 5.465 & \textbf{2.871} & 0.862 & 0.022 \\
        
        Ours + LDI & 0.469 & 2.546 & 5.479 & 2.542 & 0.863 & \textbf{0.016} \\
                
        \method{} (ours w/ 3DGS) & 0.443 & \textbf{3.305} & \textbf{5.673} & 2.772 & \textbf{0.869} & 0.025 \\
        \hline
    
    \end{tabular}
    }
    \label{tab:ablation}
        % \vspace{-0.1in}
\end{table*}

\edit{Moreover, we numerically evaluate our depth estimation against other techniques, on 30 randomly selected scenes from the real-world Stanford2D3D dataset~\cite{armeni2017joint}. For this evaluation, we use Absolute Relative Error (AbsRel), Root Mean Squared Error (RMSE), and three percentage metrics, i.e., the percentages of pixels where the ratio ($\delta$) between the estimated and ground truth depth is smaller than $1.25$, $1.25^2$, and $1.25^3$.}

\subsection{Wide-Image Reconstruction Comparisons}

Table~\ref{table:pano} numerically compares \PanoMethod{} against vanilla progressive inpainting using our base inpainting diffusion model, L-MAGIC~\cite{cai2024magic}, SyncDiffusion~\cite{shih20203d}, and MultiDiffusion~\cite{bar2023multidiffusion}. Note that, since these images are not as wide as cylindrical panoramas, we perform the optimization for 15 iterations instead of 20. As seen, our method produces better results than all the other approaches across nearly all metrics. More importantly, the images generated by \PanoMethod{} show better consistency in terms of both style and content, demonstrating the effectiveness of our optimization strategy. 

In Fig.~\ref{fig:pano-comparison}, we show visual comparison against the other approaches. As seen, progressive inpainting generates results with noticeable seams and strong inconsistency. L-Magic which works based on progressive inpainting, gradually changes the style of the image closer to the two sides. Similarly SyncDiffusion and MultiDiffusion produce results that are not consistent with the input images. Note that the walkway in center of Multidiffusion's result does not align with the surrounding regions. In contrast, \PanoMethod{} can generate coherent and seamless wide-images that are significantly better than other approaches. 
% In addition, given the same input image, \PanoMethod{} can synthesize diverse and coherent results across multiple runs, as is shown in Fig.~\ref{fig:pano_diverse}.

\subsection{Panorama Depth Estimation}
\edit{In Table~\ref{tab:depth}, we numerically evaluate the performance of \DepthMethod{} with two baseline depth estimation methods: Depth Anything V2~\cite{yang2024depthv2} (``DA V2 + ours'') and MoGe~\cite{wang2025moge} (``MoGe + ours''), a recently introduced (concurrent) single-image depth estimation method. As seen, our approach with MoGe produces better results than DA V2. We also report the metrics for MoGe with its own patch-wise depth estimation and blending approach (``MoGe''). Comparing ``MoGe'' with ``MoGe + ours'' demonstrates that \DepthMethod{} achieves better performance across most metrics. For our 3D scene reconstruction experiments, we adopt DA V2 as the base estimator, since it performs reliably across diverse settings (indoor, outdoor, synthetic) and integrates well with our pipeline. Nonetheless, our approach is agnostic to the choice of depth estimator and can be combined with MoGe or future methods, potentially yielding further improvements.}

\subsection{3D Scene Reconstruction Comparisons}

We show numerical comparisons of our \method{} against LucidDreamer~\cite{chung2023luciddreamer} and WonderJourney~\cite{yu2024wonderjourney} in Table~\ref{table:3d}. We use the official code released by the authors, and utilize the same training cameras as ours for a fair comparison. While our image quality and aesthetic scores are slightly worse than WonderJourney, our consistency scores are significantly better. This is because their novel view images are often reasonable when viewed one image at a time, however, different novel view images differ in style and thus are not consistent. Our approach, on the other hand, produces results that are consistent across all views.

We further show visual comparisons against the other methods in Fig.~\ref{fig:3d-comparison}. LucidDreamer and WonderJourney produce results with seams and inconsistent style across the two views shown here. In contrast, \method{} can synthesize consistent and seamless scene at both novel views. Please refer to our supplementary materials for video comparison and more visual results.

\subsection{Novel View Synthesis Ablations}
\begin{figure*}
    \centering
    \includegraphics[width=0.9\linewidth]{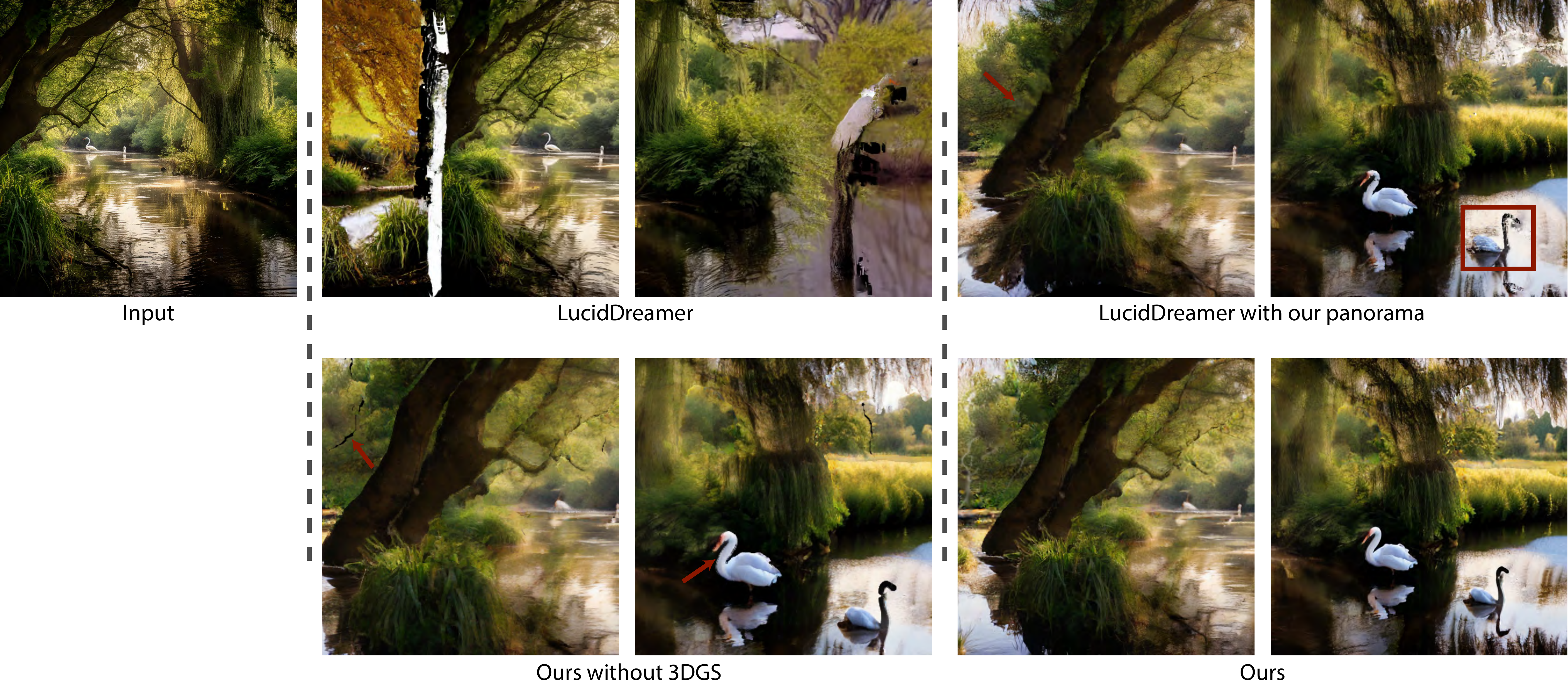}
    \vspace{-0.1in}
    \caption{\edit{We highlight the effect of various components of our novel view synthesis pipeline. Using our consistent panorama in progressive techniques like LucidDreamer~\cite{chung2023luciddreamer} improves the texture quality around input image boundaries. However, the rendered texture still contains artifacts due to inconsistent scene depth estimation. Moreover, using the intermediate layered depth image (LDI) representation instead of 3DGS (Ours) leads to subpar rendered images with with artifacts around the depth layer discontinuities.}}
    \label{fig:nvs_ablation}
\end{figure*}

\edit{We evaluate the effectiveness of different components in our 3D reconstruction pipeline on novel view synthesis task, both visually (Fig~\ref{fig:nvs_ablation}) and numerically (Table~\ref{tab:ablation}). While incorporating our coherent panorama improves LucidDreamer’s performance, their generated novel views still contain artifacts due to inconsistent depth estimation. Moreover, we show the impact of 3DGS optimization by comparing our results against LDI rendering. As seen in Fig.~\ref{fig:nvs_ablation}, LDI's rendering contains artifacts around the depth layer discontinuities. These are significantly reduced with our 3DGS optimization partly due to our use of the depth-based novel view loss.}

\section{Conclusion, Limitation, and Future Work}

In conclusion, we have presented a novel method for generating 360\textdegree{} 3D scenes from a single input image. Our approach first generates a panoramic image along with its corresponding depth map. After inpainting occluded regions, these images are used to optimize a 3DGS representation from which novel views can be rendered. To create a coherent and globally consistent panorama, we frame the task as an optimization problem with two terms, solving it effectively through an alternating minimization strategy. Additionally, we pose the problem of estimating panorama depth using an existing monocular depth estimation method as an optimization and address it with alternating minimization. Extensive experiments show that our approach outperforms state-of-the-art methods in both panorama generation and reconstructed 3D scenes. \edit{We note that although our main focus has been 3D scene reconstruction, our proposed components, \PanoMethod{}  and \DepthMethod{}, are general and have potential applications in related areas such as wide image generation and high-resolution depth prediction.}

\edit{While our method demonstrates clear advantages over prior work, it also has some limitations.} First, for our approach to generate appropriate panoramas, like all existing methods, the input image must have a mostly horizontal horizon. Additionally, our approach only reconstructs the front of objects, limiting our ability to capture the areas behind them. In the future, it would be interesting to combine our approach with existing projection-based approaches to address this limitation. \edit{Moreover, in some cases the generated textures can appear slightly blurry, particularly near the panorama edges. This results from our optimization process, which propagates content outward from the center to maintain contextual consistency, but occasionally at the cost of sharpness. Despite this, our results remain significantly better than those of existing methods. Lastly, while our approach already enables some level of scene exploration, a promising future direction is to extend this capability by using our 360\textdegree{} representation as a scaffold for state-of-the-art video generation models, enabling a more immersive scene exploration.}

\begin{acks}
The project was funded by Leia Inc. (contract \#415290). Portions of this research were conducted with the advanced computing resources provided by Texas A\&M High Performance Research Computing.
\end{acks}

% \input{X_supp}

% Bibliography
\bibliographystyle{ACM-Reference-Format}
\bibliography{main}
\clearpage
% \appendix

\clearpage
% This command starts a new two-column page, 
% but puts the content in brackets [] across the top of both columns.
\twocolumn[{%
    \begin{center}
        \Large \textbf{PanoDreamer: Optimization-Based Single Image to 360 3D Scene With Diffusion} \\
        \vspace{10pt}
        \Large \textbf{Supplementary Material}
        \vspace{20pt} % Add space between title and text
    \end{center}
}]

% --- Reset Counters for Supplement  ---
\setcounter{figure}{0}
\setcounter{table}{0}
\setcounter{equation}{0}
\setcounter{page}{1}
\renewcommand{\thefigure}{S\arabic{figure}}
\renewcommand{\thetable}{S\arabic{table}}
\renewcommand{\theequation}{S\arabic{equation}}

\begin{figure*}
    \centering
    \includegraphics[width=\linewidth]{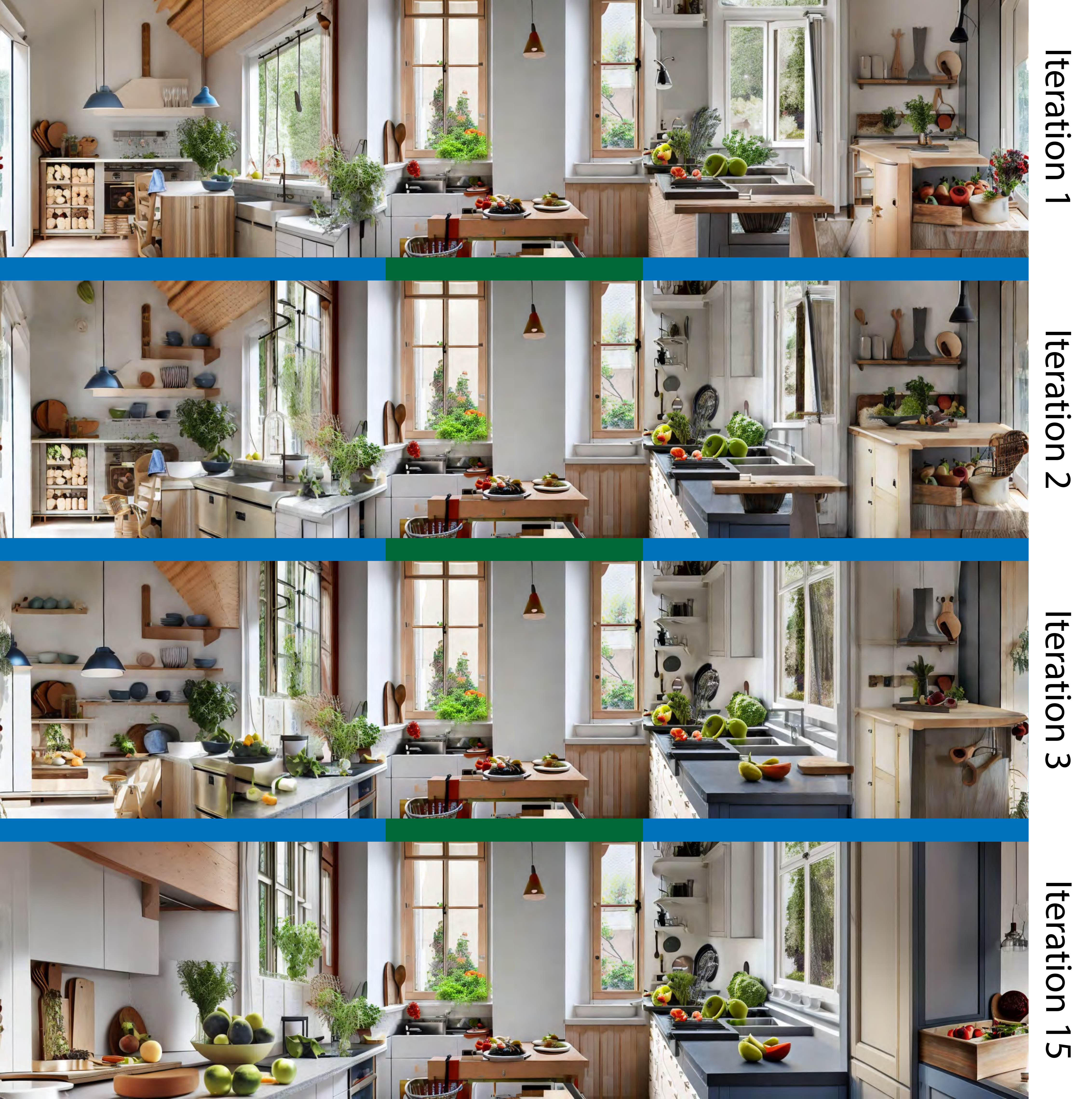}
    \vspace{-0.2in}
    \caption{We show the results of \PanoMethod{} during different iterations of the optimization.}
    \vspace{-0.2in}
    \label{fig:pano_iterations}
\end{figure*}

In this supplementary material, we present additional supporting and result figures to further validate and illustrate our findings.

\begin{figure*}
    \centering
    \includegraphics[width=\linewidth]{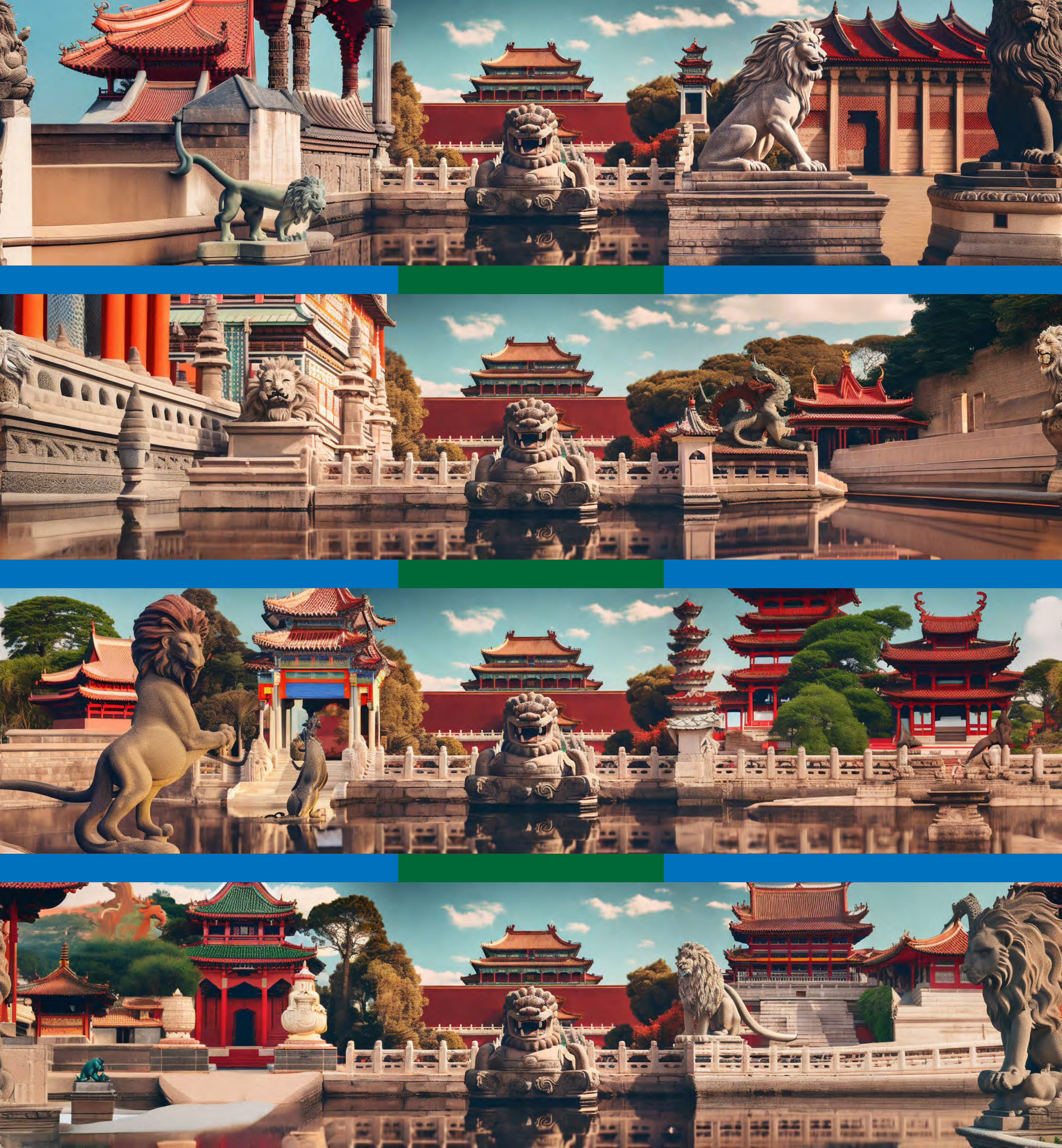}
    \vspace{-0.2in}
    \caption{We show the results of our approach on the same input image across multiple runs. As shown, our approach produces diverse yet consistent results.}
    \label{fig:pano_diverse}
    \vspace{-0.2in}
\end{figure*}

\section{Context Propagation}
First, Fig.~\ref{fig:pano_iterations} illustrates the denoised outputs obtained across different optimization iterations. As shown, with an increasing number of iterations, the central input context progressively propagates outward, ultimately converging to a consistent final result.

\section{Diversity}
As illustrated in Fig.~\ref{fig:pano_diverse}, our approach successfully generates diverse, high-quality results for different random initializations.

\section{Timing and Implementation Details}
\edit{All experiments are conducted on an NVIDIA RTX A5000 GPU. Our method requires 100 minutes per scene on average, compared to 65 minutes for WonderJourney~\cite{yu2024wonderjourney}, with panorama generation constituting the primary bottleneck (4 minutes 40 seconds per iteration; 93 minutes for 20 iterations). Computational efficiency was not the focus of this work as the method is built upon MultiDiffusion~\cite{bar2023multidiffusion}. Since the introduction of MultiDiffusion, however, several approaches such as SpotDiffusion~\cite{frolov2024spotdiffusion} have achieved up to a 10x speedup, which could potentially be directly integrated into our framework. Further acceleration can be obtained by replacing the current Stable Diffusion model (50 sampling steps) with faster variants that require as few as 4–8 steps (e.g., few-steps distilled models). Moreover, because all panorama windows (crops) are independent, the method can be parallelized across multiple GPUs to substantially reduce runtime.}

\section{MVDiffusion}
MVDiffusion~\cite{deng2023mv} is a recent diffusion-based approach designed to generate panoramic images conditioned on input views. However, since their model is trained on indoor panoramic data, it tends to overfit, resulting in poor generalization and diminished performance on out-of-distribution scenes, as shown in Fig.~\ref{fig:mvdiffusion}.

\section{Additional Results}
\begin{figure*}
    \centering
    \includegraphics[width=\linewidth]{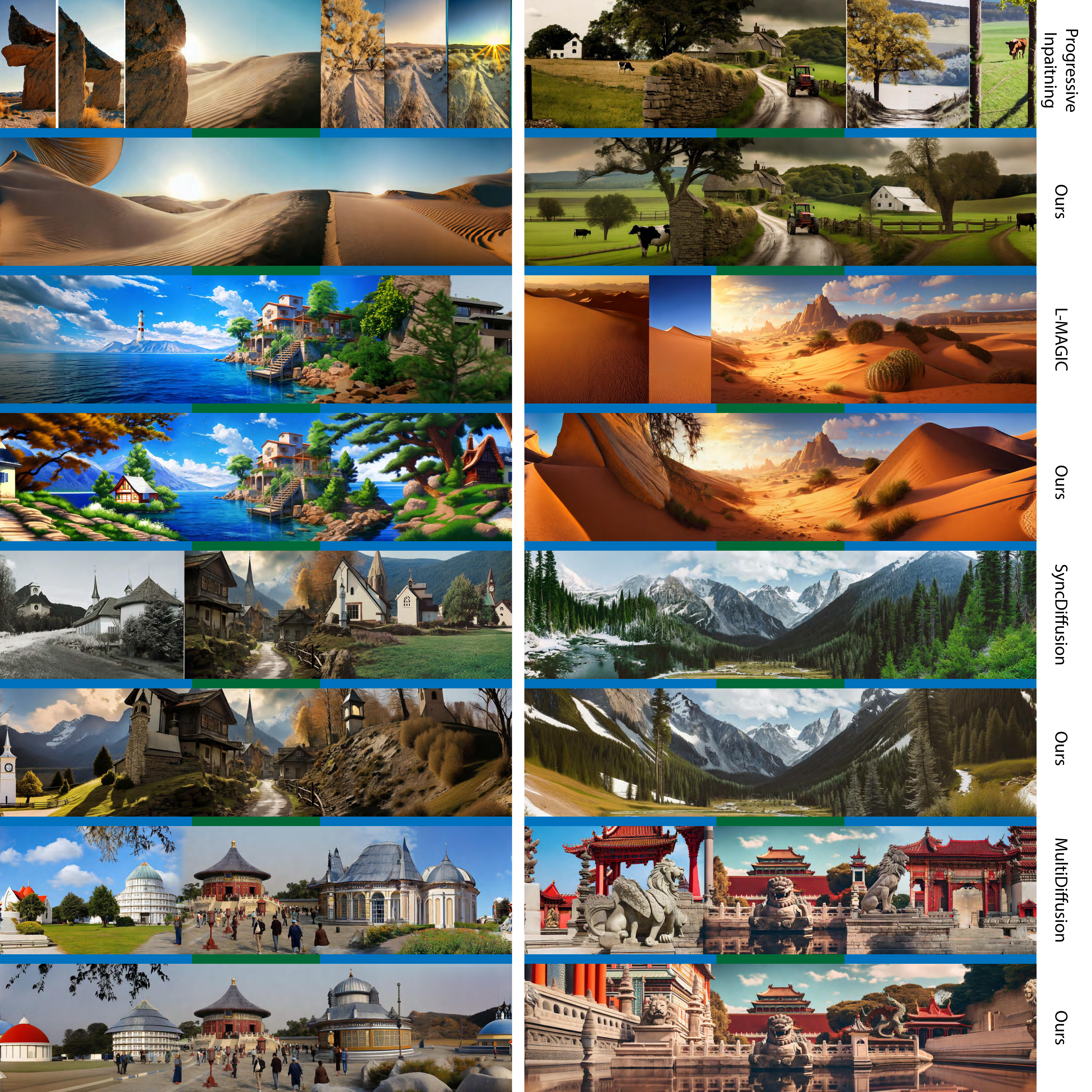}
    \caption{We compare the wide-images generated by \PanoMethod{} (Ours) with those from other methods. Other approaches often result in sharp discontinuities and contextual inconsistencies.}
    \label{fig:rebuttal_pano}
\end{figure*}

\begin{figure*}
    \centering
    \includegraphics[width=\linewidth]{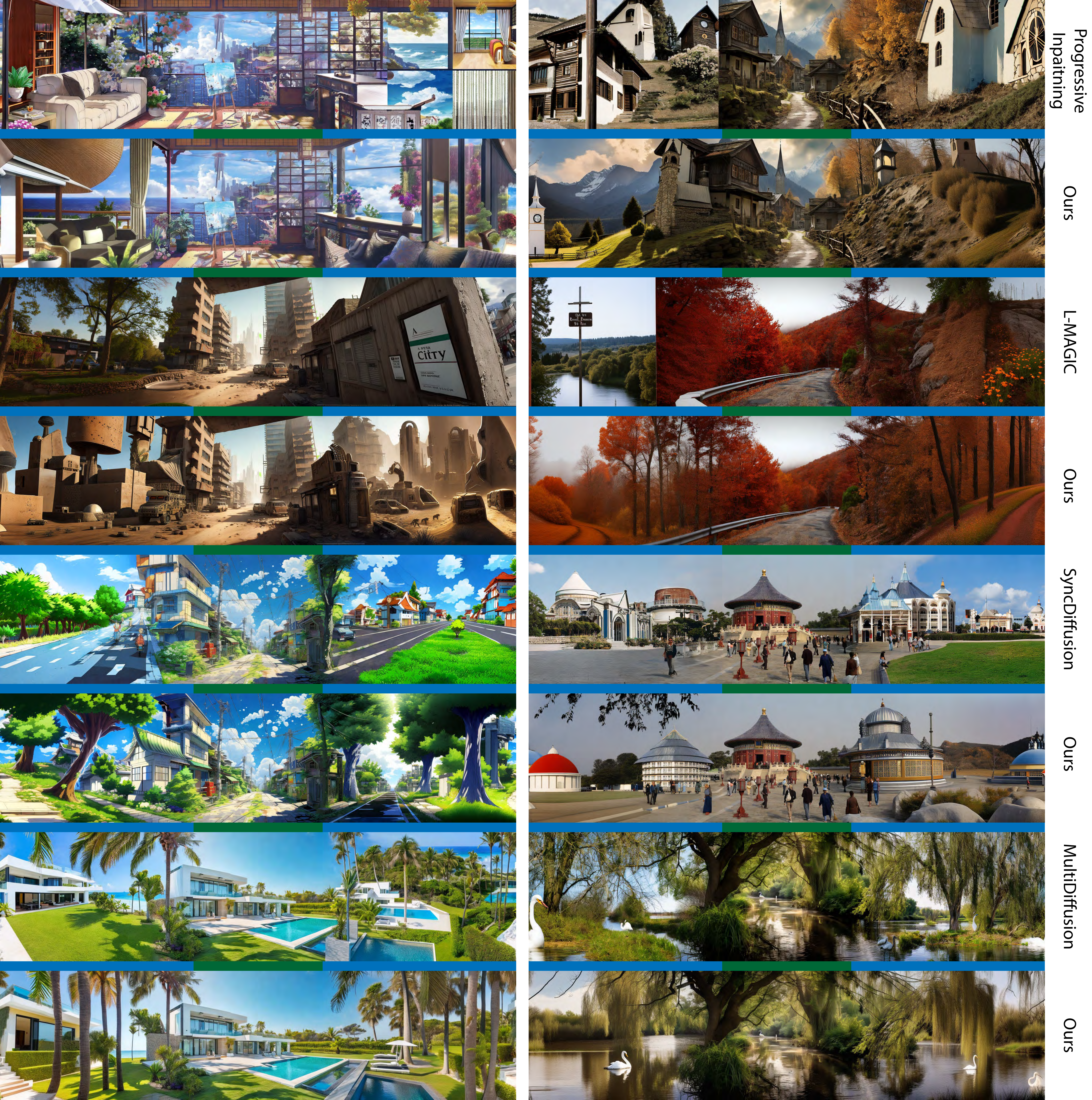}
    \caption{We compare the wide-images generated by \PanoMethod{} (Ours) with those from other methods. Other approaches often result in sharp discontinuities and contextual inconsistencies.}
    \label{fig:rebuttal_pano1}
\end{figure*}

We present additional qualitative comparisons between our method and recent state-of-the-art wide-image generation approaches~\cite{bar2023multidiffusion, lee2023syncdiffusion, cai2024magic} in Fig.\ref{fig:rebuttal_pano} and Fig.\ref{fig:rebuttal_pano1}. As illustrated, \PanoMethod{} (Ours) consistently generates more coherent and seamless panoramas, significantly outperforming existing methods.

We also present further depth comparisons with the state-of-the-art depth estimators, Depth-Anything V2~\cite{yang2024depthv2} (DA V2) \edit{and MoGe~\cite{wang2025moge} (patch-wise panorama depth estimator)}, in Fig.~\ref{fig:rebuttal_depth} and Fig.~\ref{fig:rebuttal_depth1}. As shown, our method generates depth maps with greater detail and improved consistency, particularly around panorama boundaries (left corners). We highlight prominent artifacts in DA V2's and MoGe's results using white arrows. \edit{Similar to our approach, MoGe also computes patch-wise depth and blends them to obtain panoramic depth. However, it generates slightly blurry results and can contain scene scale inconsistency in some local patches.}

\begin{figure*}
    \centering
    \includegraphics[width=\linewidth]{figures/mvdiffusion.pdf}
    \caption{MVDiffusion~\cite{deng2023mv} is a single image panorama generation approach. Since their model is trained on indoor panoramic data, it tends to overfit, resulting in poor generalization and diminished performance on out-of-distribution scenes.}
    \label{fig:mvdiffusion}
\end{figure*}

\begin{figure*}
    \centering
    \includegraphics[width=0.75\linewidth]{figures/rebuttal_depth_comp_final.pdf}
    \caption{Our method generates depth maps with greater detail and improved consistency, particularly around panorama boundaries (left corners). We highlight prominent artifacts in DA V2 and MoGe's results using white arrows.}
    \label{fig:rebuttal_depth}
\end{figure*}

\begin{figure*}
    \centering
    \includegraphics[width=0.75\linewidth]{figures/rebuttal_depth_comp1_final.pdf}
    \caption{Our method generates depth maps with greater detail and improved consistency, particularly around panorama boundaries (left corners). We highlight prominent artifacts in DA V2 and MoGe's results using white arrows.}
    \label{fig:rebuttal_depth1}
\end{figure*}

\end{document}